\documentclass{article}
\usepackage[affil-it]{authblk}
\usepackage{graphicx}
\usepackage[space]{grffile}
\usepackage{latexsym}
\usepackage{textcomp}
\usepackage{longtable}
\usepackage{times}
\usepackage{multirow,booktabs}
\usepackage{amsfonts,amsmath,amssymb}
\usepackage{url}
\usepackage{hyperref}
\hypersetup{colorlinks=false,pdfborder={0 0 0}}
\usepackage[capitalise]{cleveref}
\usepackage[english]{babel}
\usepackage{lipsum}
\usepackage{fancyhdr}
\usepackage[usenames,dvipsnames]{color} % for color text
\usepackage{comment} % To comment out blocks of text
\usepackage{float} % To put figures/tables exactly where I want them
\usepackage{tablefootnote} % To add footnotes below tables
\usepackage{scrextend} % to use \footref: multiple reference to the same table footnote
\usepackage{pbox} % to have new line inside table cells
\usepackage{gensymb} % to have the \degree symbol
\usepackage{epstopdf}
\usepackage{subfig}
\usepackage[linesnumbered, vlined]{algorithm2e}
\makeatletter
\def\blx@maxline{77}
\makeatother
\usepackage[flushleft]{threeparttable}
\usepackage[acronym,nonumberlist]{glossaries} 
\usepackage{glossary-mcols}  
\usepackage{glossary-longragged}
\usepackage{tabu}
\usepackage{pdflscape} % to add table in landscape
\usepackage{luca} % some macros that Luca is used to using... 

\newglossary[slg]{symbolslist}{syi}{syg}{List of Symbols}
\newglossaryentry{e}{name=\emph{e},description={Error factor}, user1={}, type=symbolslist, sort=error}

\newacronym{SNS}{SNS}{social network site}
\newacronym{WWW}{WWW}{World Wide Web}
\newacronym{LDA}{LDA}{Latent Dirichlet Allocation}

\newglossarystyle{acronymstyle}{
\renewenvironment{theglossary}
 {\begin{longtabu} to \linewidth {p{0.2\linewidth}p{0.8\linewidth}}}%
 {\end{longtabu}}%% Set the table?s header: title row
}

\newcommand{\email}[1]{{\small \tt #1}}

\newglossarystyle{symbolstyle}{%
 {\begin{longtabu} to \linewidth {p{0.2\linewidth}p{0.55\linewidth}p{0.25\linewidth}}}%
 {\end{longtabu}}%% Set the table?s header: title row
}

\makeglossaries
\usepackage{lineno}
\begin{document}

% To add the title
\title{\emph{Some Like it Hoax}:\\ Automated Fake News Detection in Social Networks}
\author[1]{Eugenio~Tacchini}
\author[2]{Gabriele~Ballarin}
\author[3]{Marco~L.~Della~Vedova}
\author[4]{Stefano~Moret}
\author[5]{Luca~de~Alfaro}
\affil[1]{Universit\`a Cattolica, Piacenza, Italy. \email{eugenio.tacchini@unicatt.it}}
\affil[2]{Independent researcher. \email{gabriele.ballarin@gmail.com}}
\affil[3]{Universit\`a Cattolica, Brescia, Italy. \email{marco.dellavedova@unicatt.it}}
\affil[4]{\'Ecole Polytechnique F\'ed\'erale de Lausanne, Switzerland. \email{moret.stefano@gmail.com}}
\affil[5]{Department of Computer Science, UC Santa Cruz, CA, USA. \email{luca@ucsc.edu}}
\date{\normalsize Technical Report UCSC-SOE-17-05 \\ 
School of Engineering, University of California, Santa Cruz}

\maketitle

% To add content to the title page
%\setcounter{tocdepth}{2}
%\tableofcontents

% This ensures that only the main file is compiled
% !TEX root = Hoax_paper.tex

% Abstract & Intro - Stefano

\begin{abstract}
In recent years, the reliability of information on the Internet has emerged as a crucial issue of modern society. 
Social network sites (SNSs) have revolutionized the way in which information is spread by allowing users to freely share content. 
As a consequence, SNSs are also increasingly used as vectors for the diffusion of misinformation and hoaxes. 
The amount of disseminated information and the rapidity of its diffusion make it practically impossible to assess reliability in a timely manner, highlighting the need for automatic hoax detection systems.

As a contribution towards this objective, we show that Facebook posts can be classified with high accuracy as hoaxes or non-hoaxes on the basis of the users who ``liked'' them. 
We present two classification techniques, one based on logistic regression, the other on a novel adaptation of boolean crowdsourcing algorithms. 
On a dataset consisting of 15,500 Facebook posts and 909,236 users, we obtain classification accuracies exceeding 99\% even when the training set contains less than 1\% of the posts. 
We further show that our techniques are robust: 
they work even when we restrict our attention to the users who like both hoax and non-hoax posts.
These results suggest that mapping the diffusion pattern of information can be a useful component of automatic hoax detection systems.
\end{abstract}

\section{Introduction}

% Motivation: web + SN revolutionize informaiton -> hoaxes -> fast diffusion + high quantity -> need of automatic system
% add some example of hoaxes?
The \gls{WWW} has revolutionized the way in which information is disseminated. 
In particular, \glspl{SNS} are platforms where content can be freely shared, enabling users to actively participate to  - and, possibly, influence - information diffusion processes. 
%This phenomenon is often referred to as \emph{collective intelligence} \textcolor{red}{[Ref?]}. 
As a consequence, \glspl{SNS} are also increasingly used as vectors for the dissemination of spam~\cite{heymann_fighting_2007}, conspiracy theories and \emph{hoaxes}, i.e. intentionally crafted fake information. 
This recently led to the emphatic definition of our current times as the \emph{age of misinformation}~\cite{bessi_science_2015}. 
A significant share of hoaxes on \glspl{SNS} diffuses rapidly, with a peak in the first 2 hours~\cite{del_vicario_spreading_2016}. 
This finding, together with the high amount of shared content, highlights the need of automatic online hoax detection systems~\cite{hernandez_first_2002}.

% Literature review: previous efforts in automatic hoax detection systems
In the literature, various approaches have been proposed for automatic hoax detection, covering quite heterogeneous applications.
% emails: Vukovic, Petkovic, Yevseyevaa, Ishak. Webpages: Sharifi
Historically, one of the first applications has been hoax detection in e-mail messages and webpages.
In the context of scam e-mail detection,  spamassassin uses keyword-based methods with logistic regression \cite{mason2002filtering}; Petkovi\'c et al.~\cite{petkovic_e-mail_2005} and Ishak et al.~\cite{ishak_distance-based_2012} proposed the use of distance-based methods; Vukovi\'c et al.~\cite{vukovic_intelligent_2009} applied neural network and advanced text processing; Yevseyeva et al.~\cite{yevseyeva_optimising_2013} used evolutionary algorithms for the development of anti-spam filters. Sharifi et al.~\cite{sharifi_detection_2011} applied logistic regression to automatically detect scam on webpages, reaching an accuracy of 98\%.

% Moving towards social applications: reputation -> Wikipedia
The concepts of trust and reputation~\cite{Mui:2002:CMT:820745.821158,dellarocas2003digitization} can be adopted for hoax detection in applications with a dominant social component. Metrics and algorithms for this purpose have been proposed by Golbeck and Hendler~\cite{golbeck_accuracy_2004}. Adler and de Alfaro~\cite{Adler:2007:CRS:1242572.1242608} developed a content-driven user reputation system for Wikipedia, allowing to predict the quality of new contributions. 
The detection of Wikipedia hoaxes has been addressed e.g.\ in \cite{potthast2008automatic,adler2011wikipedia,Kumar:2016:DWI:2872427.2883085}. 
%
% Social Network: Twitter
More recently, automatic hoax detection in \glspl{SNS} has gained increasing interest. As an example, Chen et al.~\cite{chen_scam_2014} developed a semi-supervised scam detector for Twitter based on self-learning and clustering analysis, while Ito et al.~\cite{ito_assessment_2015} proposed the use of \gls{LDA} to assess the credibility of tweets.

The key idea behind our work, which constitutes its main novelty, is that hoaxes can be identified with great accuracy on the basis of the users that interact with them. 
In particular, focusing on Facebook, we answer the following research question: \emph{Can a hoax be identified based on the users who ``liked" it?}
We consider a dataset consisting of 15,500 posts and 909,236 users; the posts originate from pages that deal with either scientific topics or with conspiracies and fake scientific news \cite{bessi_science_2015}. 
We propose two classification techniques. 
One consists in applying logistic regression, considering the user interaction with posts as features. 
The other technique consists in a novel adaptation of boolean label crowdsourcing techniques to a setting where a training set is available, but no prior assumption on users being mostly reliable can be made. 

The proposed techniques yield an accuracy exceeding 99\% even for training sets consisting of less of 1\% of posts.
These results are obtained in spite of the fact that the communities of users participating in the scientific and conspiracy pages overlap.
% We also show that our methods work when the classifier is trained on a subset of the Facebook pages (and hence, communities) that constitute our complete dataset.
%
Our main contributions, in summary, are: 
\emph{i}) the proposal of a novel way to identify hoaxes on \glspl{SNS} based on the users who interacted with them rather than their content;
\emph{ii}) an improved version of the harmonic crowdsourcing method, suited to hoax detection in \glspl{SNS}; 
\emph{iii}) the application on Facebook and, in particular, on a representative dataset obtained from the literature.

The code we developed for this paper is available from \url{https://github.com/gabll/some-like-it-hoax}.

\section{Dataset}
\label{sec:dataset}

Our dataset consists in all the public posts and posts' likes of a list of selected Facebook pages during the second semester of 2016:
from Jul. 1st, 2016 to Dec. 31st, 2016.
We collected the data by means of the Facebook Graph API%
\footnote{See \url{https://developers.facebook.com/docs/graph-api}. We used version 2.6.}
on Jan. 27th, 2017.
% , so the considered posts and likes are those available at that moment.

We based our selection of pages on~\cite{bessi_science_2015}.
In that work, the authors present a list of Facebook pages divided into two categories: scientific news sources vs.\ conspiracy news sources.
We assume all posts from scientific pages to be reliable, i.e. ``non-hoaxes'', and all posts from conspiracy pages to be ``hoaxes''.
Among the 73 pages listed in~\cite{bessi_science_2015}, we limited our analysis to the top 20 pages of both categories.
It is worth noting that at the time of data collection, not all the pages were still available:
some of them had been deleted in the meantime, or were no longer publicly accessible.
We note also that the actual posts comprising our dataset are distinct from those originally included in the dataset of \cite{bessi_science_2015}, as we performed our data collection in a different, and more recent, period.

The resulting dataset, the so-called \emph{complete dataset}, is composed of 15,500 posts from 32 pages (14 conspiracy and 18 scientific), with more than 2,300,00 likes by 900,000+ users (\cref{fig-datasets}).
Among posts, 8,923 (57.6\%) are hoaxes and 6,577 (42.4\%) non-hoaxes.

\begin{figure}[t]
  \centering
  \subfloat[]{\includegraphics[width=.49\columnwidth]{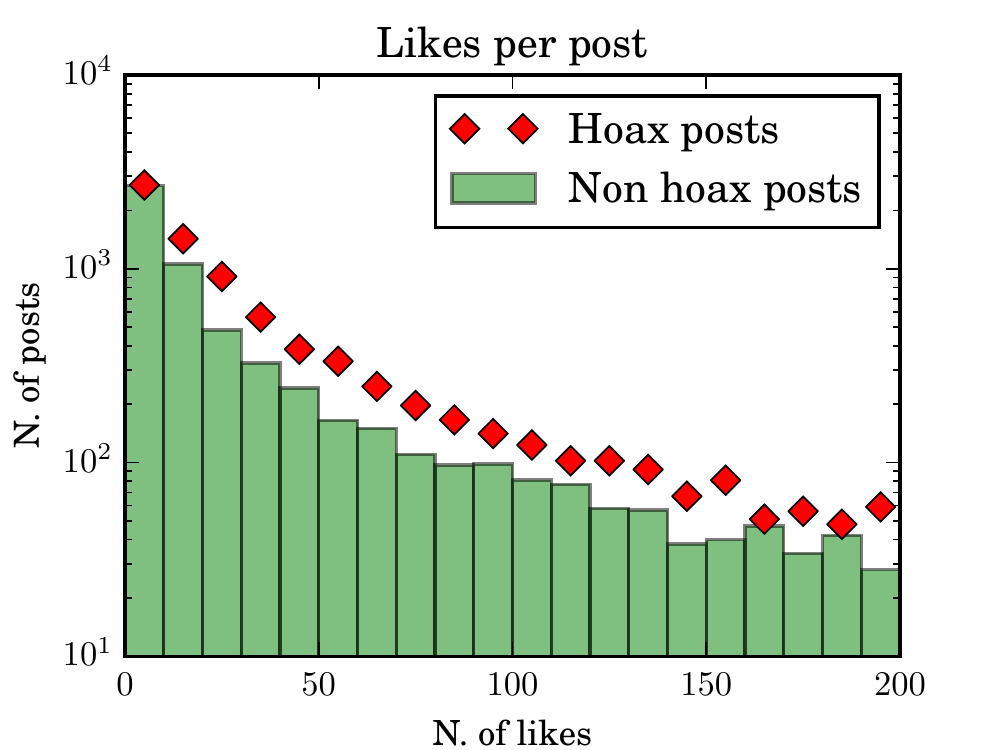}}
  \subfloat[]{\includegraphics[width=.49\columnwidth]{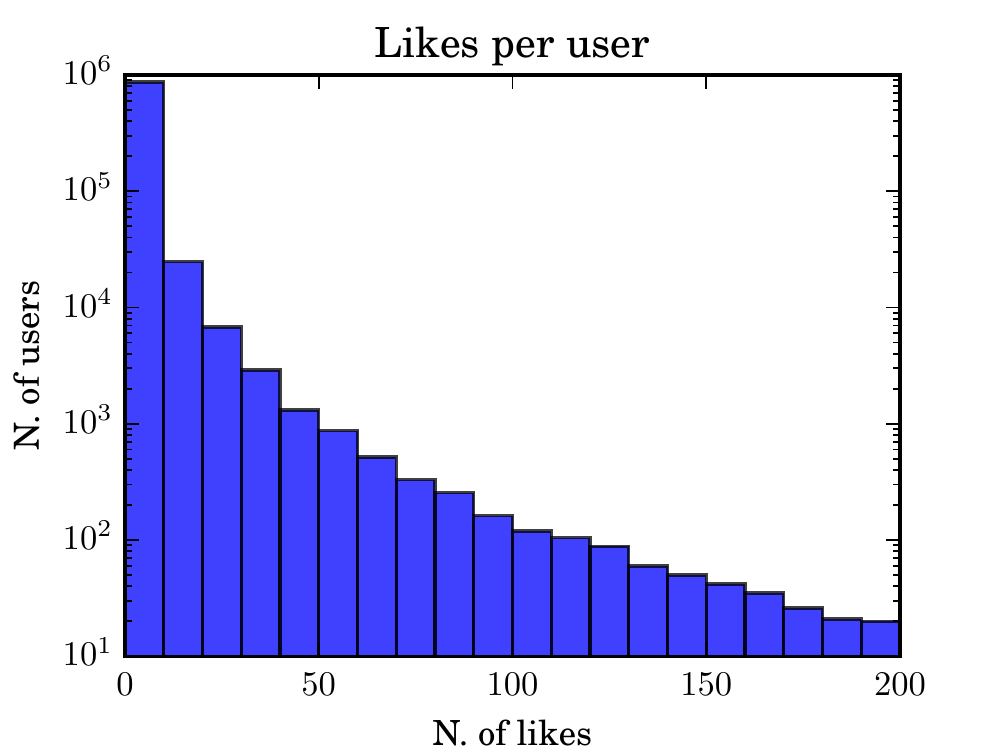}}
  \caption{Likes per post (a) and likes per user (b) histograms for the dataset. Plots are represented with semi-logarithmic scale.
    \label{fig:likes_hist}}
\end{figure}

As a first observation, the distribution of the number of likes per post is exponential-like, as attested by the histograms in \cref{fig:likes_hist}~(a); the majority of the posts have few likes.
% In particular, there are 5,747 posts (37\%) with less than 10 likes.
Hoax posts have, on average, more likes than non-hoax posts.
In particular, some figures about the number of likes per post are: 
average, 204.5~(for hoax post) vs. 84.0~(non-hoax);
median, 22~(hoax) vs. 14~(non-hoax);
maximum, 121,491~(hoax) vs. 13,608~(non-hoax).

\begin{figure}[t]
  \centering
  \subfloat[]{\includegraphics[width=.49\columnwidth]{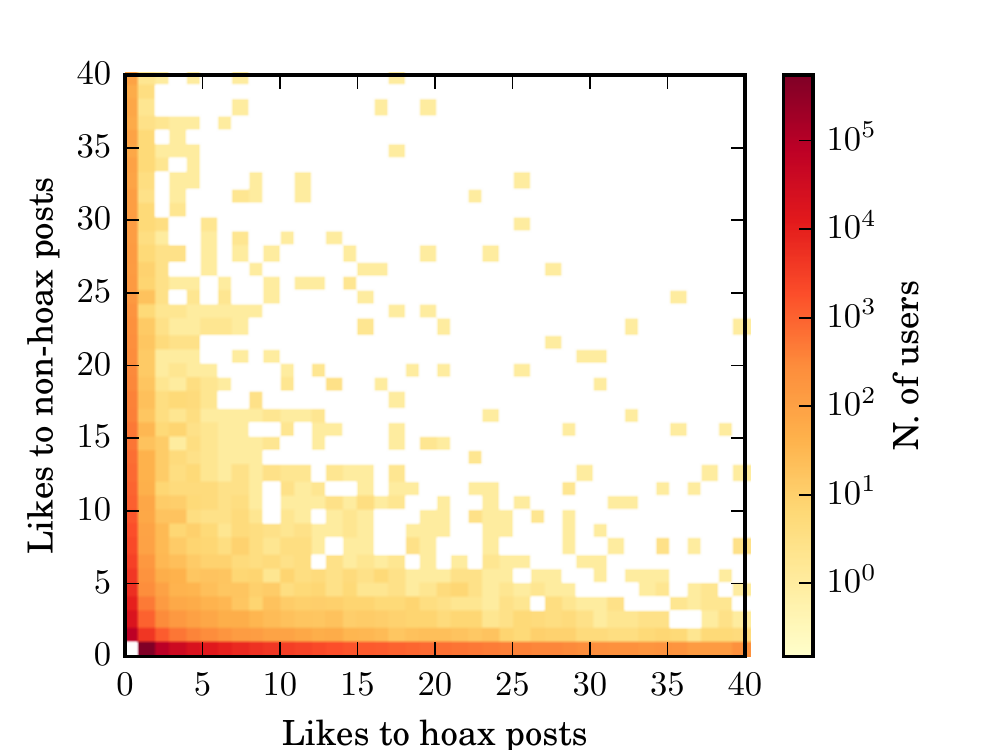}}
%   \subfloat[]{\includegraphics[width=.49\columnwidth]{./fig/bothusers_hist}}
  \subfloat[]{\includegraphics[width=.49\columnwidth]{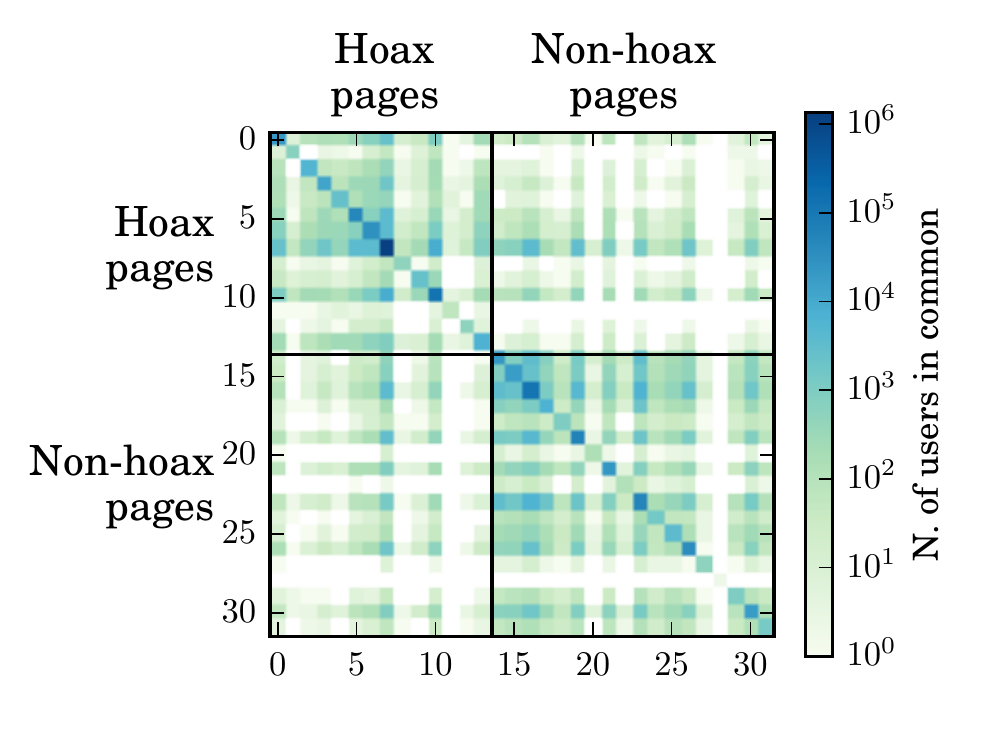}}
  \caption{Users characterization: hoax vs. non-hoax likes per user heat-map (a) and 
    users in common between pages (b)
%   hoax likes proportion histogram for mixed-users with 10+ likes (b).
    \label{fig:users}}
\end{figure}

A second observation is related to the number of likes per user:
once again, \cref{fig:likes_hist}~(b) shows an exponential-like distribution.
The majority of the users appears in the dataset with one single like (629,146 users, 69.2\%), 
while the maximum number of likes by a user is 1,028.
Users can be divided into three categories based on what they liked:
\textit{i}) those who liked hoax posts only, \textit{ii}) those who liked non-hoax posts only, and \textit{iii}) those who liked at least one post belonging to a hoax page, and one belonging to a non-hoax page.
\cref{fig:users}~(a) shows that, despite a high polarization, there are many users in the mixed category: among users with at least 2 likes, 209,280 (74.7\%) liked hoax post only, 56,671 (20.3\%) liked non-hoax post only, and 14,139 (5.0\%) are in the mixed category.
This latter category gives rise to the \emph{intersection dataset}, which consists only of the users who liked {\em both\/} hoax and non-hoax posts, and of the posts these users liked. 
The intersection dataset was introduced to study the performance of our methods for communities of users that are not strongly polarized towards hoax or non-hoax posts, as will be discussed in \cref{sec:results}.
The composition of the intersection dataset is summarized in \cref{fig-datasets}.

% The proportion of mixed-likers increases if we do not consider users with too few likes: for example, among the 38,711 users with 10 likes or more, 2,936 (7.9\%) are in the mixed category.
% For these latter users (mixed category, 10+ likes), the proportion of likes to hoax posts referred to the total number of likes is represented in \cref{fig:users}~(b):
% the effect of the polarization is evident from the U-shaped histogram; nevertheless, there are 604 ``unpolarized'' users with 10+ likes, characterized by a hoax likes ratio between 0.25 and 0.75.

\begin{table}[t]
  \centering
  \begin{tabular}{l|r|r|r}
  & N. posts & \quad N. users & \quad N. likes \\ 
  \hline
  Complete & 15,500 & 909,236 & 2,376,776 \\
  Intersection \quad & 10,520 & 14,139 & 117,641 \\
  \end{tabular}
  \medskip
  \caption{Composition of complete and intersection datasets.}
  \label{fig-datasets}
\end{table}

A third observation concerns the relation between pages, measured by the number of users that pages have in common:
given each pair of pages, we study how many users liked at least one post from one page and one post from the other page.
\cref{fig:users}~(b) shows the result as a symmetric matrix: each page vs. each other page.
Color intensity displays that hoax pages have more users in common with other hoax pages (up-left part, which appears darker) than with non-hoax pages (up-right and bottom-left).
The same applies to non-hoax pages (bottom-right).
Nevertheless, the figure shows that the communities gravitating around hoax and non-hoax pages share many common users (as evidenced also from the composition of the intersection dataset).
% Nevertheless, the figure also indicates a certain polarization, as there are more common users across pages of similar classification, than across pages of different classification.

\section{Algorithmic Classification of Posts}

Our goal is to classify posts into {\em hoax\/} and {\em non-hoax\/} posts. 
According to the analysis of social media sharing by \cite{del_vicario_spreading_2016}, ``users tend to aggregate in communities of interest, which causes reinforcement and fosters confirmation bias, segregation, and polarization'', and ``users mostly tend to select and share content according to a specific narrative and to ignore the rest.''
This suggests that the set of users who like a post should be highly indicative of the nature of the post. 
We present two approaches, one based on logistic regression, the other based on boolean crowdsourcing algorithms.  

\subsection{Classification via logistic regression}

We formulate the post classification problem as a supervised learning, binary classification problem. 
We consider a set of posts $I$ and a set of users $U$. 
Each post $i \in I$ has an associated set of features $\set{x_{iu} \mid u \in U}$, where $x_{iu} = 1$ if $u$ liked post $i$, and $x_{iu} = 0$ otherwise. 
We classify the posts on the basis of their features, that is, on the basis of which users liked them. 

To perform the classification, we use a logistic regression model. 
The logistic regression model learns a weight $w_u$ for each user $u \in U$; the probability $p_i$ that a post $i$ is non-hoax is then given by $p_i = 1/(1 + e^{-y_i})$, where $y_i = \sum_{u \in U} x_{iu} w_u$.
Intuitively, $w_u > 0$ (resp.\ $w_u < 0$) indicates that $u$ likes mostly non-hoax (resp.\ hoax) posts.

We chose logistic regression for two reasons.
First, logistic regression is well suited to problems with a very large, and uniform, set of features. 
In our case, we have about a million features (users) in our dataset, but a real application would involve up to hundreds of millions of users. 
Second, our logistic regression setting enjoys a {\em non-interference\/} property with respect to unrelated set of users that facilitates learning, and is appealing on conceptual grounds. 
Specifically, assume that the set of users and posts are partitioned into disjoint subsets $U = U_1 \union U_2$, $I = I_1 \union I_2$, so that users in $U_k$ like only posts in $I_k$, for $k = 1, 2$. 
This situation can arise, for instance, when there are two populations of users and posts in different languages, or simply when two topics are very unrelated. 
In such a setting, it is equivalent to train a single model, or to train separately two models, one for $I_1$, $U_1$, one for $I_2$, $U_2$, and then take their ``union''.  
This because the weights $w_u$ for $u \in U_{3-k}$ do not matter for classifying posts in $I_k$, $k = 1,2$, since the features $x_{iu}$ with $i \in I_k$ and $u \in U_{3-k}$  are all zero.
In other words, models for unrelated communities do not interfere: if we learn a model for $I_1$, $U_1$, we do not need to revise the model once the community $I_2$, $U_2$ is discovered: all we need to do is learn a model of this second community, and use it jointly with the first.

\subsection{Classification via harmonic boolean label crowdsourcing}

The weak aspect of logistic regression is that it does not transfer information across users who liked some of the same posts. 
In particular, if the training set does not contain any post liked by a user $u$, then logistic regression will not be able to learn anything about $u$, and $w_u$ will be undetermined. 
Thus, posts that are only liked by users not in the training set cannot be classified. 
As an alternative approach, we propose to perform the hoax/non-hoax classification using algorithms derived from crowdsourcing, and precisely, from the {\em boolean label crowdsourcing\/} (BLC) problem.

In the BLC problem, users provide True/False labels for posts, indicating for instance whether a post is vandalism, or whether it violates community guidelines. 
The BLC problem consists in computing the consensus labels from the user input \cite{karger_iterative_2011,liu_variational_2012,de_alfaro_reliable_2015}.
We model liking a post as voting True on that post. 

Our setting differs from standard BLC in one important respect.
Standard BLC algorithms do not use a learning set: rather, they assume that people are more likely to tell the truth than to lie.
The algorithms compare what people say, correct for the effect of the liars, and reconstruct a consensus truth \cite{karger_iterative_2011,de_alfaro_reliable_2015}.
In our setting, we cannot assume that users are more likely to tell the truth, that is, like preferentially non-hoax posts.
Indeed, hoax articles may well have more ``likes'' than non-hoax ones. 
Rather, we will rely on a learning set of posts for which the ground truth is known. 

We present here an adaptation of the {\em harmonic\/} algorithm of \cite{de_alfaro_reliable_2015} to a setting with a learning set of posts. 
We chose the harmonic algorithm because it is computationally efficient, can cope with large datasets, and it offers good accuracy in practice, as evidenced in \cite{de_alfaro_reliable_2015}. 
Furthermore, while the harmonic algorithm can be adapted to the presence of a learning set, it is less obvious how to do so for some of the other algorithms, such as those of \cite{karger_iterative_2011}.

We represent the dataset as a bipartite graph $(I \union U, L)$, where $L \subs I \times U$ is the set of likes.
We denote by $\partial i = \set{u \mid (i, u) \in L}$ and $\partial u = \set{i \mid (i, u) \in L}$ the 1-neighborhoods of a post $i \in I$ and user $u \in U$, respectively.

The harmonic algorithm maintains for each node $v \in I \union U$ two non-negative parameters $\alpha_v$, $\beta_v$. 
These parameters define a beta distribution: intuitively, for a user $u$, $\alpha_u - 1$ represents the number of times we have seen the user like a non-hoax post, and $\beta_u - 1$ represents the number of times we have seen the user like a hoax post.
For a post $i$, $\alpha_i - 1$ represents the number of non-hoax votes it has received, and $\beta_i - 1$ represents the number of hoax votes it has received. 
For each node $v$, let $p_v = \alpha_v / (\alpha_v + \beta_v)$ be the mean of its beta distribution: for a user $u$, $p_u$ is the (average) probability that the user is truthful (likes non-hoax posts), and for a post $i$, $p_i$ is the (average) probability that $i$ is not a hoax.
Letting $q_v = 2 p_v - 1 = (\alpha_v - \beta_v) / (\alpha_v + \beta_v)$,  positive values of $q_v$ indicate propensity for non-hoax, and negative values, propensity for hoax. 

Let the training set consist of two subets $I_H, I_N \subs I$ of known hoax and non-hoax posts.
The algorithm sets $q_i := -1$  for all $i \in I_H$, and $q_i := 1$ for all $i \in I_N$; it sets $q_i = 0$ for all other posts $i \in I \setm (I_H \union I_N)$.
The algorithm then proceeds by iterative updates.
First, for each user $u \in U$, it lets: 
\begin{align} 
    \alpha_u & := A + \sum \set{q_i \mid i \in \partial u, q_i > 0}
    & \qquad
    \beta_u & :=  B - \sum \set{q_i \mid i \in \partial u, q_i < 0}
    \nonumber \\
    q_u & := (\alpha_u - \beta_u) / (\alpha_u + \beta_u)
    \eqpun . & & \label{eq-user-update}
\end{align}
The positive constants $A$, $B$ determine the amount of evidence needed to sway the algorithm towards believing that a user likes hoax or non-hoax posts: the higher the values of $A$ and $B$, the more evidence will be required. 
After some experimentation, we settled on the values $A = 5.01$ and $B = 5$, corresponding to a very weak a-priori preference of users for non-hoax posts.
This corresponds to needing about 5 ``likes'' from known good (resp\ bad) users to reach a 2:1 probability ratio in favor of non-hoax (resp.\ hoax), which seems intuitively reasonable. 
The algorithm then updates the values for each post $i \in I \setm (I_H \union I_N)$ by:
\begin{align} 
    \alpha_i & := A' + \sum \set{q_u \mid u \in \partial i, q_u > 0}
    \qquad
    \beta_i & :=  B' - \sum \set{q_u \mid u \in \partial i, q_u < 0}
    \nonumber \\
    q_i & := (\alpha_i - \beta_i) / (\alpha_i + \beta_i)
    \eqpun . & & \label{eq-post-update}
\end{align}
We choose $A' = B' = 5$, thus adopting a symmetrical a-priori for items being hoax vs.\ non-hoax. 
The updates (\ref{eq-user-update})--(\ref{eq-post-update}) are performed iteratively; while they could be performed until a fixpoint is reached, we just perform them 5 times, as further updates do not yield increased accuracy.
Finally, we classify a post $i$ as hoax if $q_i < 0$, and as non-hoax otherwise. 

The harmonic algorithm satisfies the {\em non-interference\/} property described for logistic regression, since information is only propagated along graph edges that correspond to {\em ``likes''}.

The harmonic algorithm is able to propagate information from posts where the ground truth is known, to posts that are connected by common users. 
In the first iteration, the users who liked mostly hoax (resp.\ non-hoax) posts will see their $\beta$ (resp.\ $\alpha$) coefficient increase, and thus their preferences will be characterized. 
In the next iteration, the user preferences will be reflected on post beliefs, and these post beliefs will subsequently be used to infer the preferences of more users, and so on. 
We will see how the ability to transfer information will allow the harmonic algorithm to reach high levels of accuracy even starting from small training sets.

\section{Results}
\label{sec:results}

% I think it's always nice to give a bit of a preview to the results. 
% Some version of this, more precise and numerical, should go into the intro.
We characterize the performance of the logistic regression and harmonic BLC algorithm via two sets of experiments. 
The first set of experiments measures the accuracy of the algorithms as a function of the number of posts available as training set. 
Since the training set can be produced, in general, only via a laborious process of manual post inspection, these results tell us how much do we need to invest in manual labeling, to reap the benefits of automated classification.
The second set of experiments measures how much information our learning is able to transfer from one set of pages to another.  
As the community of Facebook users is organized around pages, these experiments shed light on how much what we learn from one community can be transferred to another, via the shared users among communities.

\subsection{Accuracy of classification vs.\ training set size}

\paragraph{Cross-validation analysis.}
We performed a standard cross-validation analysis of logistic regression and of the harmonic algorithm for BLC. 
The cross-validation was performed by dividing the posts in the dataset into 80\% training and 20\% testing, and performing a 5-fold cross-validation analysis. 
Both approaches performed remarkably well, with accuracies exceeding 99\% for logistic regression and 99.4\% for the harmonic algorithm. 

\paragraph{Accuracy vs.\ training set size.}
Cross-validation is not the most insightful evaluation of our algorithms. 
In classifying news posts as hoax or non-hoax, there is a cost involved in creating the training set, as it may be necessary to examine each post individually. 
The interesting question is not the level of accuracy we can reach when we know the ground truth for 80\% of the posts, but rather, how large a training set do we need in order to reach a certain level of accuracy. 
In order to be able to scale up to the size of social network information sharing, our approaches need to be able to produce an accurate classification relying on a small fraction of posts of known class. 

To better understand this point, it helps to contrast the situation for standard ML settings, versus our post-classification problem.
In standard ML settings, the set of features is chosen in advance, and  the model that is developed from the 80\% of data in the training set is expected to be useful for all future data, and not merely the 20\% that constitutes the evaluation set.
Thus, cross-validation provides a measure of performance for any future data.
In contrast, in our setting the ``features'' consist in the users that liked the posts. 
The larger the set of posts we consider, the larger the set of users that might have interacted with them; 
we cannot assume that the model developed from 80\% of our data will be valid for any set of future posts to be classified. 
Rather, the interesting question is, how many posts do we need to randomly select and classify, in order to be able to automatically classify all others? 

% % Here we give alternative figures.  Please comment out the ones we don't use; don't delete them.

% \begin{figure}
% \centering 
% \subfloat[]{
%     \includegraphics[scale=0.42]{accuracy_vs_fraction_all_s.pdf}
%     }
% \subfloat[]{
%     \includegraphics[scale=0.42]{accuracy_vs_fraction_int_s.pdf}
%     }    
% \caption{Accuracy (fraction of correctly classified posts) of the logistic and harmonic BLC algorithms on (a) the complete dataset and (b) the intersection dataset, as a function of the fraction of posts in the  training set.  The data is the average of 50 runs; the error bars denote the standard deviation of the accuracy value of each run.}
% \label{fig-perc}
% \end{figure}

% % Another variant. 
% \begin{figure}
% \centering 
%     \includegraphics[scale=0.5]{accuracy_vs_fraction_all_b.pdf} \\
%     \includegraphics[scale=0.5]{accuracy_vs_fraction_int_b.pdf}
% \caption{Accuracy of the logistic and harmonic BLC algorithms on the complete and intersection datasets, as a function of the fraction of posts in the  training set.  The data is the average of 50 runs; the error bars denote the standard deviation of the accuracy value of each run.}
% \label{fig-perc-bigger}
% \end{figure}

% All together now.
\begin{figure}[t]
\centering
\includegraphics[scale=0.6]{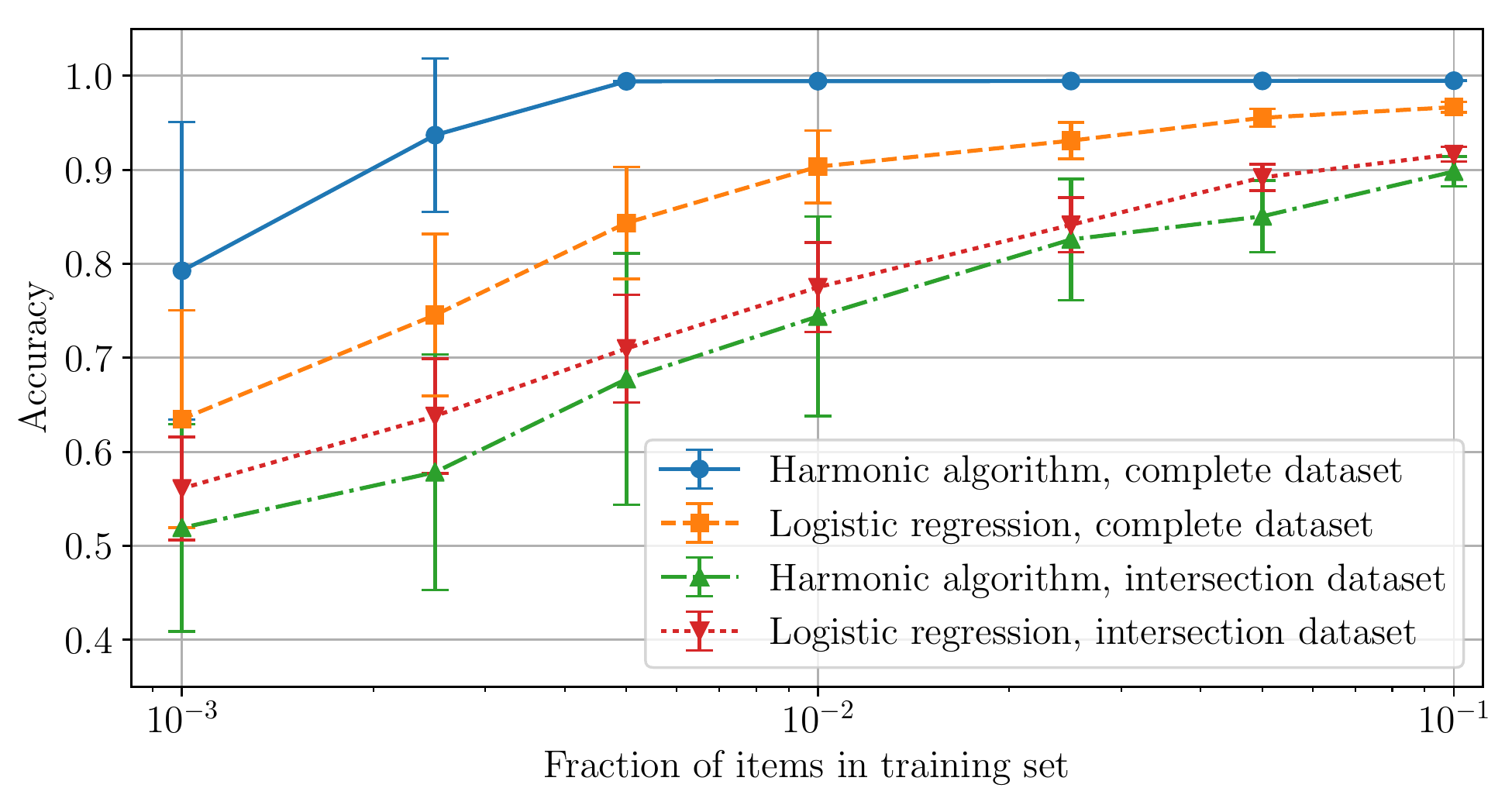}
\caption{Accuracy of the logistic and harmonic BLC algorithms on the complete and intersection datasets, as a function of the fraction of posts in the  training set.  The data is the average of 50 runs; the error bars denote the standard deviation of the accuracy value of each run.}
\label{fig-perc-joint}
\end{figure}

We report the classification accuracy both for the complete dataset, and for the intersection dataset (see \cref{sec:dataset}).

In \cref{fig-perc-joint}, we report the accuracy our methods as a function of the size of the training set. 
In the figure, the classification accuracy reported for each training set size is the average of 50 runs. 
In each run, we select randomly a subset of posts to serve as training set, and we measure the classification accuracy on all other posts.
The error bars in the figure denote the standard deviation of the classification accuracy of each run.
Thus, the error bars provide an indication of run-to-run variablity (how much the accuracy varies with the particular training set), rather than of the precision in measuring the average accuracy. 
The standard deviation with which the average accuracy is known is about seven times smaller.

For the complete dataset, the harmonic BLC algorithm is the superior one. 
As long as the training set contains at least 0.5\% of the posts, or about 80 posts, the accuracy exceeds 99.4\%.
For even lower training set sizes the accuracy decreases, but it is still about 80\% for a training set consisting of 0.1\% of posts, or about 15 posts. 
Logistic regression is somewhat inferior, but still yields accuracy above 90\% for training sets consisting of only 1\% of the posts. 

On the intersection dataset, on the other hand, the logistic regression approach is the superior one. 
While the differences between the logistic regression and harmonic BLC algorithms is not large, the performance of logistic regression starts at 91.6\% for a training set consisting of 10\% of posts, and degrades towards 56\% for a training set consisting of 0.1\% of posts, maintaining a performance margin of 3--4\% over harmonic BLC. 

Generally, these results indicate that harmonic BLC is more efficient at transfering information across the dataset. 
Its inferior performance for the intersection dataset may be explained by the fact that the artificial construction of the intersection dataset biases towards the transfer of erroneous information.
Most users have only a few likes (see Figure~\ref{fig:likes_hist}). 
The intersection dataset filters out all users who liked only one post, and of the users who liked two posts, the intersection dataset filters out all those who liked two posts of the same hoax/non-hoax class. 
As a consequence, the intersection dataset heavily over-samples ``straddling'' users who like exactly two posts, one hoax, one not; these straddling users constitute 32\% of the users in the intersection dataset.
When the two posts liked by a straddling user belong one to the training, one to the evaluation dataset, the straddling user contributes in the wrong direction to the classification of the post in the evaluation set.

\subsection{Cross-page learning}

As the community of Facebook users naturally revolves around common interests and pages, an interesting question concerns whether what we learn from one community of users on one page transfers to other pages.
In order to answer this question, we test our classifiers on posts related to pages that they have not seen during the training phase.
To this end, we perform two experiments in which the set of pages from which we learn, and those on which we test, are disjoint. 
In the first experiment, {\em one-page-out,} we select in turn each page, and we place all its posts in the testing set; the posts belonging to all other pages are in the training set.
In the second experiment, {\em half-pages-out,} we perform 50 runs. 
In each run, we randomly select a set consisting of half of the pages in the dataset, and we place the posts belonging to those pages in the testing set, and all others in the training set. 
The results are reported in \cref{fig-leave-pages}. 

The results clearly indicate that the harmonic BLC algorithm is the superior one for transferring information across pages, achieving essentially perfect accuracy in both one-page-out and half-page-out experiments. 
Surprisingly, for harmonic BLC, the performance is slightly superior in the half-pages-out than in the one-page-out experiments. 
This is due to the fact that for one page the performance is only 87.3\%; the performance for all other pages is always above 97.2\%, and is 100\% for 23 pages in the dataset. 
The poor performance on one particular page drags down the average for one-page-out, compared to half-pages-out where better-performing pages ameliorate the average. 
%
% We note that the two algorithms have different weak points. 
% The lowest performance of 87.3\% for harmonic BLC occurs over a page consisting of 211 posts; for that same page, logistic regression yields 99.5\% accuracy. 
% Conversely, logistic regression achieves minimum accuracy of 35.8\% (indicating that classification is right only about $1/3$ of the time!) for another page, consisting of 522 posts; for that page, harmonic BLC provides 96\% accuracy. 
% This suggests the existence of algorithms that combine the strenghts of logistic regression and harmonic BLC, a topic worthy of future investigation, in particular as more data becomes available.
% Overall, harmonic BLC is the algorithm that offers more consistent performance.

\begin{table}[t]
\centering
\begin{tabular}{l|r|r|r|r|}
 & \multicolumn{2}{c|}{One-page-out} 
 & \multicolumn{2}{c|}{Half-pages-out} \\
  & Avg accuracy & Stdev & Avg accuracy & Stdev \\ \hline
 Logistic regression \quad & 0.794 & 0.303 & 0.716 & 0.143 \\
 Harmonic BLC & 0.991 & 0.023 & 0.993 & 0.002 
\end{tabular}
\medskip
\caption{Accuracy (fraction of correctly classified posts) when leaving one page out, and when leaving out half of the pages, from the training set. For one-page-out, we report average and standard deviation obtained by leaving out each page in turn.  For half-pages-out, we report average and standard deviation of the accuracy obtained in 50 runs.}
\label{fig-leave-pages}
\end{table}

\section{Conclusions}

The high accuracy achieved by both logistic regression and the harmonic BLC algorithm confirm our basic hypothesis: the set of users that interacts with news posts in social network sites can be used to predict whether posts are hoaxes. 

We presented two techniques for exploiting this information: one based on logistic regression, the other on boolean label crowdsourcing (BLC). 
Both algorithms provide good performance, with the harmonic BLC algorithm providing accuracy above 99\% even when trained over sets of posts consisting of 0.5\% of the full dataset (or about 80 posts).
This suggests that the algorithms can scale up to the size of entire social networks, while requiring only a modest amount of manual classification. 

We also analyzed the extent to which our performance depends on the community of users naturally aggregating around pages of similar content. 
We showed that the harmonic BLC algorithm can transfer information across pages: even when only half of the pages are represented in the training set, the performance is above 99\%. 
Even on the ``intersection dataset'', consisting of only users who liked {\em both\/} hoax and non-hoax posts, our methods achieve performance of 90\%, albeit requiring for this a training set consisting of 10\% of the posts;
this produces evidence that our approach might work even when applied to communities of users that are not strongly polarized towards scientific vs.\ conspiracy pages.
We note that the intersection dataset is a borderline example that does not occur in the communities we studied. 
Together, these results seem to indicate that the techniques proposed may be sufficiently robust for an extensive application in a real-world scenario.

\bibliographystyle{plain} 
\bibliography{Hoaxers}

\end{document}